\documentclass[nohyperref]{article}

\usepackage{microtype}
\usepackage{graphicx}
\usepackage{booktabs} 

\usepackage{microtype}
\usepackage{graphicx}
\usepackage{booktabs} 

\usepackage{natbib}
\usepackage{amssymb, amsmath,amsthm}
\usepackage{amsfonts}
\usepackage{algorithm}
\usepackage{algorithmic}
\usepackage{paralist}
\usepackage{multirow}
\usepackage{dsfont}
\usepackage{mathrsfs}
\usepackage{times}
\usepackage{ulem}
\usepackage{wrapfig}

\usepackage{url,enumerate}
\usepackage{color,xcolor}
\usepackage{makeidx}  
\usepackage{amsmath,amssymb}
\usepackage{mathtools}
\usepackage[small, compact]{titlesec}
\usepackage{xspace}
\usepackage{epstopdf}
\usepackage{cite}
\usepackage{mathrsfs}
\usepackage{times}
\usepackage{enumerate}
\usepackage{color}
\usepackage{graphicx,epsfig}
\usepackage{amsmath,amssymb,xspace}
\usepackage{xurl}
\usepackage{hyperref}
\usepackage{bm}
\usepackage{upgreek}
\usepackage{cleveref}
\usepackage{multirow}
\usepackage{ulem}
\usepackage{cancel}
\usepackage{subcaption}

\usepackage{hyperref}

\usepackage[accepted]{icml2023}

\usepackage{amsmath}
\usepackage{amssymb}
\usepackage{mathtools}
\usepackage{amsthm}

\theoremstyle{plain}

\theoremstyle{definition}

\theoremstyle{remark}

\usepackage[textsize=tiny]{todonotes}

\icmltitlerunning{Diffusion-Generative Multi-Fidelity Learning for Physical Simulation}

\newcommand{\ours}{{{DGMF}}\xspace}

\renewcommand{\d}{{\rm d}}  
\newcommand{\e}{{\bf e}}
\newcommand{\f}{{\bf f}}

\newcommand{\h}{{\bf h}}

\newcommand{\m}{{\bf m}}

\newcommand{\s}{{\bf s}}

\newcommand{\w}{{\bf w}}
\newcommand{\x}{{\bf x}}
\newcommand{\y}{{\bf y}}
\newcommand{\z}{{\bf z}}

\renewcommand{\H}{{\bf H}}
\newcommand{\I}{{\bf I}}

\newcommand{\N}{\mathcal{N}}  

\newcommand{\Dcal}{\mathcal{D}}

\newcommand{\Xcal}{\mathcal{X}}
\newcommand{\Ycal}{\mathcal{Y}}

\newcommand{\Lcal}{\mathcal{L}}

\newcommand{\Hcal}{\mathcal{H}}

\newcommand{\W}{{\bf W}}

\newcommand{\btheta}{{\boldsymbol{\theta}}}

\newcommand{\bSigma}{\boldsymbol{\Sigma}}

\newcommand{\bmu}{\boldsymbol{\mu}}

\newcommand{\0}{{\bf 0}}

\newcommand{\ben}{\begin{enumerate}}
\newcommand{\een}{\end{enumerate}}

\newcommand{\ie}{{\textit{i.e.,}}\xspace}
\newcommand{\eg}{{\textit{e.g.,}}\xspace}
\newcommand{\etc}{{\textit{etc.}}\xspace}
\newcommand{\EE}{\mathbb{E}}

\newcommand{\cmt}[1]{}

\begin{document}

\twocolumn[
\icmltitle{Diffusion-Generative Multi-Fidelity Learning for Physical Simulation}



\icmlsetsymbol{equal}{*}

\begin{icmlauthorlist}
\icmlauthor{Zheng Wang}{yyy}
\icmlauthor{Shibo Li}{yyy}
\icmlauthor{Shikai Fang}{yyy}
\icmlauthor{Shandian Zhe}{yyy}
\end{icmlauthorlist}

\icmlaffiliation{yyy}{Kahlert School of Computing, University of Utah, Salt Lake City, United States}

\icmlcorrespondingauthor{Shandian Zhe}{zhe@cs.utah.edu}

\icmlkeywords{Machine Learning, ICML}

\vskip 0.3in
]



\printAffiliationsAndNotice{\icmlEqualContribution} 

\begin{abstract}
Multi-fidelity surrogate learning is important for physical simulation related applications in that it avoids running numerical solvers from scratch, which is known to be costly, and it uses multi-fidelity examples for training and greatly reduces the cost of data collection. Despite the variety of existing methods, they all build a model to map the input parameters outright to the solution output. Inspired by the recent breakthrough in generative models, we take an alternative view and consider the solution output as generated from random noises. We develop a diffusion-generative multi-fidelity (\ours) learning method based on stochastic differential equations (SDE), where the generation is a continuous denoising process. We propose a conditional score model to control the solution generation by the input parameters and the fidelity. By conditioning on additional inputs (temporal or spacial variables), our model can efficiently learn and predict multi-dimensional solution arrays.  Our method naturally unifies discrete and continuous fidelity modeling. The advantage of our method in several typical applications shows a promising new direction for multi-fidelity learning. 
\end{abstract}

\section{Introduction}
\vspace{-0.05in}
Physical simulation is crucial in scientific and engineering applications.  The major task of physical simulation is to solve partial differential equations (PDEs) at domains of interest. In practice, running numerical solvers is often costly. An efficient strategy is to learn a data-driven surrogate model~\citep{kennedy2000predicting,conti2010bayesian}.  Given the PDE parameters (including those in the equation, initial and/or boundary conditions),  one can use the surrogate model to predict  the high-dimensional solution output (at a specified grid), \eg via a forward pass in a neural network. This is often much cheaper and faster than running a numerical solver from scratch. 

Nonetheless, the training examples for the surrogate model still have to be produced by the numerical solvers, which can be very costly. To address this issue, an important strategy is  multi-fidelity learning.  High-fidelity examples are accurate yet costly to produce while low-fidelity examples are inaccurate but much cheaper to compute.  By effectively integrating information from multi-fidelity data, one can still expect to achieve good predictive performance while greatly reducing the cost of training data collection. Many multi-fidelity modeling and learning approaches have been developed, such as~\cite{perdikaris2017nonlinear,parussini2017multi,xing2021deep,wang2021multi}, many of which use an auto-regressive structure to capture the relationship among the fidelities. The recent work of~\citet{li2022infinite} considers continuous-fidelity modeling as they observe that the fidelity is often determined by the mesh spacing (or finite element length) in the numerical solvers and hence is continuous in nature. 

Despite the difference between existing works, they can all be viewed as ``likelihood'' based  approaches. That is, they directly model the complex mapping between the PDE parameters and the solution output at different fidelities, and then maximize the model likelihood  to fit the training data. In parallel, we observe that in the domain of generative modeling, which aims to generate samples of a target distribution (\eg images of a certainty type) from random noises, the recent breakthrough ---  diffusion-generative modeling~\citep{yang2022diffusion,croitoru2022diffusion} --- can largely outperform the popular likelihood-based counterparts, such as auto-regressive models~\citep{larochelle2011neural,van2016pixel} and variational auto-encoders~\citep{kingma2013auto,rezende2014stochastic}. The diffusion-generative framework uses a forward diffusion model, \eg a Markov chain~\citep{ho2020denoising}, to gradually corrupt the data (\eg an image) until the data has become random noises. It then learns a backward denoising or generation model (\eg another Markov chain) to gradually remove the noise and to recover the data.  The model essentially is to fit the noise rather than the data. It turns out that training with small steps of diffusion and denoising is more stable and powerful than directly learning a complex model of the target distribution. 

Discretized PDE solutions can be viewed as special images. Predicting PDE solutions  bears a similarity to image generation --- one most successful task of the generative models. Therefore, gradually producing the solution prediction via a similar diffusion-denoising process might also bolster the accuracy. To this end,  we develop \ours, a novel  Diffusion-Generative Multi-Fidelity learning method. The contributions of our work are summarized as follows. 
\begin{compactitem}
	\item First, based on  the recent score-based SDE framework~\citep{song2021scorebased}, we model the discretized PDE solution as generated from Gaussian white noises, and the target distribution is conditioned on the PDE parameters and the fidelity. The diffusion and denoising processes are fulfilled by a forward SDE and a reverse SDE, respectively. The reverse SDE is determined by the score of the forward SDE distribution, which is intractable. We propose a conditional model to learn the score, where the PDE parameters and fidelity are used as additional model inputs to control the solution generation. These inputs are transformed via a multi-layer perceptron, and added into each ResNet block along the contraction and expansion paths of a U-Net~\citep{ronneberger2015u}; see Fig. \ref{fig:score-model-framework}. 
	\item Second, to predict multi-dimensional solution arrays, \eg a spatial-temporal tensor, we further condition our model on temporal and/or spatial variables in the solution space to generate a 2D slice each time. In this way, we greatly reduce the model complexity and maintain the training efficiency. Predicting multiple slices in the solution can be done in parallel.
	\item Third, \ours naturally unifies the discrete and continuous fidelity modeling, where for the former, we use one-hot encoding and for the latter use the continuous fidelity value as the corresponding model input.
	\item Fourth, we show that \ours reduces the $L_2$ relative error by 55.6\%, 58\% and 45.4\% in predicting solutions of three benchmark PDEs, Poisson's, Heat and Burger's equations, respectively, as compared with the state-of-the-art (see Table \ref{tb:err-reduct}). \ours also greatly outperforms the current methods in predicting optimal topology structures and flow dynamics, with 18.1\% and 21\% improvement, respectively.  The results have demonstrated a promising new direction in multi-fidelity modeling and learning.  
\end{compactitem}

\section{Background}\label{sect:bk}
\subsection{Multi-Fidelity Learning} 
In physical simulation related applications, one often needs to compute a function with a high-dimensional output while the input is low-dimensional. For example, given a viscosity (a scalable input),  one wants to obtain the solution of the viscous Burger's equation at a $128 \times 128$ spatial-temporal grid. It is costly to numerically solve the function for every input. Hence, we seek for learning a surrogate model to predict the function values outright given the inputs. A popular method is Linear Model of Coregionalization (LMC)~\citep{journel1978mining}, which models the function output as the linear combination of a set of bases, 
\begin{align}
\f(\x) = \sum\nolimits_{k=1}^{K} z_k(\x) \h_k = \sum\nolimits_{k=1}^K \H \cdot \z(\x) \label{eq:lcm}
\end{align}
where $\H = [\h_1, \ldots, \h_K]$ are the $K$ bases and $\z(\x) = [z_1(\x), \ldots, z_K(\x)]$ the coefficient functions. The bases $\H$ are often constructed from Principled Component Analysis (PCA)~\citep{higdon2008computer}, and the coefficient functions estimated from Gaussian process (GP) regression~\citep{Rasmussen06GP}. 

We still need to run numerical solvers to prepare the training data. To enable a cost and accuracy trade-off, one can adopt different mesh spacings or finite-element sizes in the solver to produce examples at multiple fidelities. Many multi-fidelity learning methods have been developed. For example, \citet{li2020deep} introduced a set of different coefficient functions $\z_m(\x)$ and  bases $\H_m$ for each fidelity $m$, and used an auto-regressive neural network to capture the relationship between successive fidelities,  
\begin{align}
	\z_{m+1}(\x) = \text{NN}(\z_{m}(\x), \x), \;\;	\f_m(\x) = \H_m \z_m(\x)
\end{align} 
where $\f_m(\x)$ is the function output at fidelity $m$ and $\text{NN}$ a neural network.  \citet{wang2021multi} proposed a multi-fidelity extension of LCM (see \eqref{eq:lcm}), which uses a similar auto-regressive structure, but places a matrix GP prior to model each $\z_m$ as a function of both $\z_{m-1}$ and $\H_{m-1}$. 

\citet{li2022infinite} observed that the fidelity is often determined by the mesh spacing or finite-element size, which is continuous in nature. Hence, they viewed the fidelity $m$ as continuous and used a neural ordinary differential equation (ODE) to model $\z(m)$: $\frac{\partial \z}{\partial m} = \text{NN}(\z, m, \x)$, which can be viewed as a continuous extension of auto-regressive and residual modeling. The bases $\H(m)$ also vary continuously with $m$, and they model $\H(m)$ with element-wise neural ODEs or GP regression.

\subsection{Deep Generative Models}
The goal of generative models is to generate samples of a target distribution, \eg a certain type of images and audio. The target distribution is usually high-dimensional and very complicated. Given a collection of examples $\Ycal = \{\y_1, \ldots, \y_N\}$, many works use deep neural networks to estimate the density of the target distribution, such as
auto-regressive models~\citep{larochelle2011neural,germain2015made,van2016pixel},  normalization flows~\citep{dinh2014nice,dinh2016density,papamakarios2017masked}, and variational auto-encoders~\citep{kingma2013auto,rezende2014stochastic}. They can be called likelihood based approaches since they maximize the (approximate) likelihood for model estimation. The recent diffusion-generative modeling framework largely outperforms the likelihood-based approaches. The idea is to introduce a diffusion process to gradually add noises into the examples of the target distribution until they become completely random, and then we learn a denoising model to gradually remove the noises and recover those examples. For example, \citet{sohl2015deep,ho2020denoising}  modeled the diffusion process via a Markov chain, $q(\y_0, \y_1, \ldots, \y_T) = q(\y_0) \prod_{t=1}^T   \N(\y_t; \sqrt{1 - \beta_t} \y_{t-1}, \beta_t \I)$ where $q(\y_0)$ is the target distribution and $\{\beta_t\}$ is a (constant) variance schedule. When $T$ is large, we can view $q(\y_T)$ approximately as a standard Gaussian distribution. The denoising (or generation) process is modeled by another Markov chain, $p(\y_T, \y_{T-1}, \ldots, \y_0) = \N(\y_T|\0, \I) \prod_{t=1}^{T} p(\y_{t}|\y_{t-1})$, where  
$p(\y_{t-1}|\y_{t}) = \N(\y_{t-1}| \bmu_{\btheta}(\y_t, t), \bSigma_{\btheta}(\y_t, t))$,  $\bmu_{\btheta}$ is parameterized by a neural network and $\bSigma_{\btheta}$ is usually set to the covariance of  $q(\x_t|\x_{t-1}, \x_0)$, which is a constant.  Recently, \citet{song2021scorebased} used SDEs to extend the discrete Markov-chains to continuous diffusion and denoising processes, and estimated a score-based generative model. The framework is more flexible and general. 

\begin{figure*}[h]
	\centering
	\setlength\tabcolsep{0pt}
	\captionsetup[subfigure]{aboveskip=0pt,belowskip=0pt}
	\begin{tabular}[c]{c}
		\begin{subfigure}[t]{\textwidth}
			\centering
			\includegraphics[width=0.8\textwidth]{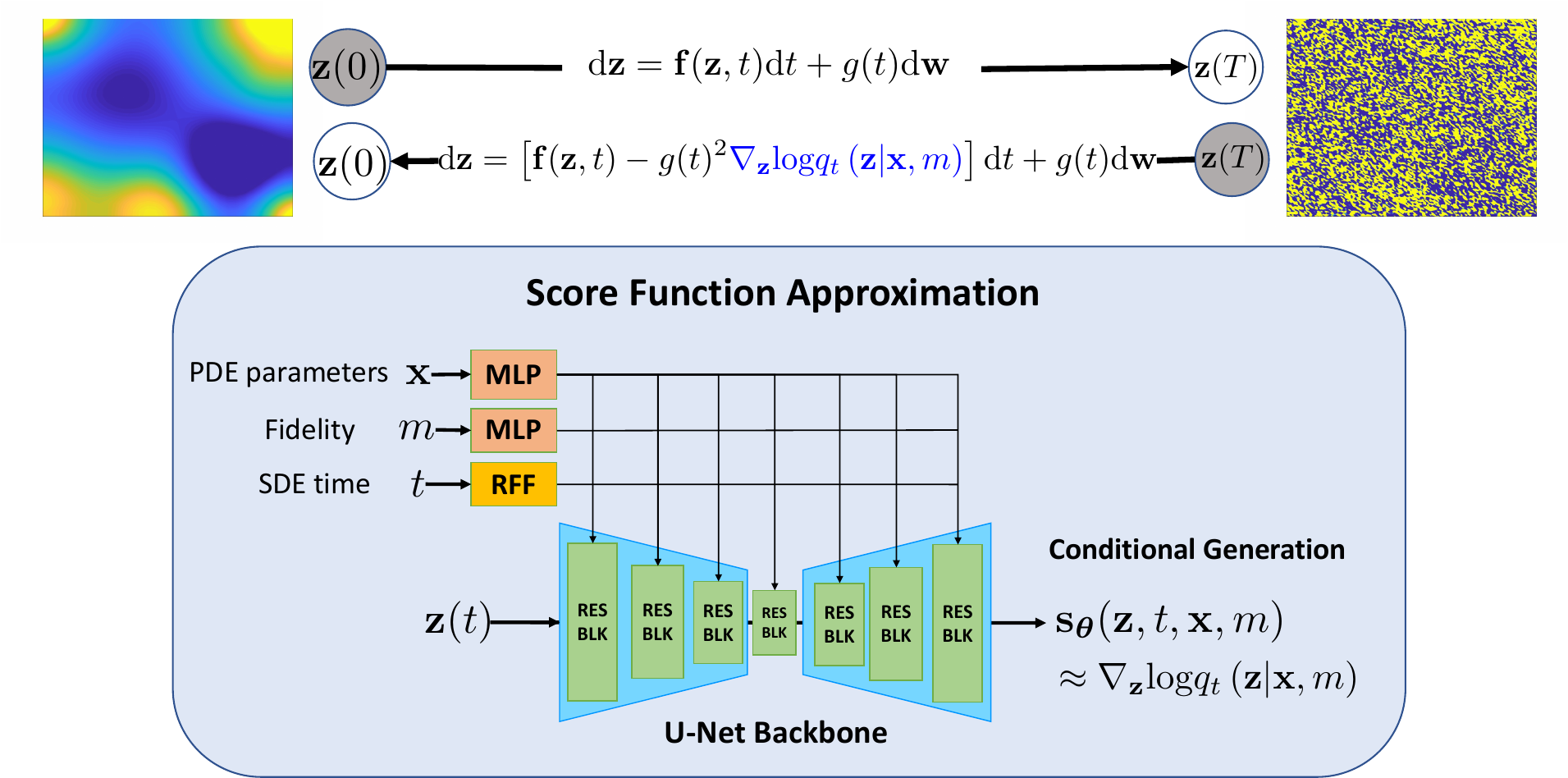}
		\end{subfigure} 
	\end{tabular}
	\caption{\small  The architecture of the diffusion-generative multi-fidelity (\ours) surrogate model.}
	\label{fig:score-model-framework}
\end{figure*}


\section{Diffusion-Generative Multi-Fidelity Surrogate Learning} 
\subsection{Model}
In this section, we propose a novel diffusion-generative multi-fidelity model for physical simulation. Suppose we have collected a set of multi-fidelity examples, $\Dcal = \{(\x_n, m_n, \bar{\y}_n)\}$, where $\x_n$ is the PDE parameters, $\bar{\y}_n$ is the (high-dimensional) discretized solution output, and $m_n$ is the fidelity. Note that $m_n$ can be discrete (\eg $m_n = 1, 2, \ldots$) or continuous, the latter corresponding to the mesh spacing or length of finite elements. We assume that given an arbitrary valid input  $\x \in \Xcal$, the solution output at fidelity $m$, which we denote by $\y_m(\x)$, follows a conditional distribution, 
\begin{align}
	\y_m(\x) \sim p(\cdot | \x, m).
\end{align}
Hence, predicting the PDE solution (at different fidelities) can be considered as generating a sample from the conditional distribution $p(\cdot | \x, m)$, for which we build a diffusion-generative model.  First, we construct a diffusion process by a forward SDE that continuously converts the solution $\y_m(\x)$ into a Gaussian white noise, 
\begin{align}
	\d \z(t) &= \f(\z, t) \d t + g(t) \d \w \notag \\
	 \z(0) & \overset{\Delta}{=}  \y_m(\x) \sim p(\cdot | \x, m) \label{eq:forward-sde}
\end{align}
where $\z(t)$ is the state at time $t$, $\w$ is the Brownian motion, and $\f$ and $g$ are drift and diffusion coefficients, respectively. For computational convenience, we can choose a simple form for $\f$ and $g$.  In our experiment, we used the variance exploding (VE) SDE~\citep{song2021scorebased}, where $\f(\z, t) = \0$, $g(t) = \sqrt{\frac{\d v(t)}{\d t}}$, and   $v(t)>0$ is a variance function. 

We denote the probability density of $\z_{m}(t)$ by $q_t(\cdot| \x, m)$. Given a large $T>0$, we can consider that after a long-time diffusion, $q_T$ has  (approximately)  become the distribution of a Gaussian white noise. If we start from samples of $\z(T) \sim q_T$ and reverse the process, we can obtain the solution sample $\z(0) \sim q_0 \overset{\Delta}{=} p(\cdot | \x, m)$. According to~\citep{anderson1982reverse}, the reversed process is the following SDE, 
\begin{align}
	\d \z(t) &= [\f(\z, t) - g(t)^2 \nabla_{\z} \log q_t(\z|\x, m)] \d t \notag \\
	&+ g(t) \d \hat{\w}, \notag \\
	\z(T) &\sim q_T(\cdot|\x, m), \label{eq:backward}
\end{align}
where $\hat{\w}$ is another Brownian motion. The reverse SDE, namely, the generation process, is determined  by the score of the forward SDE distribution, $\nabla_{\z} \log q_t(\z|\x, m)$, which is intractable and we need to estimate from data. Since the solution is determined by the PDE parameters $\x$ and fidelity $m$, the generation process should be controlled by $\x$ and $m$. Thus, we  propose a conditional score model $\s_\btheta(\z, t, \x, m)$ to estimate $\nabla_{\z} \log q_t(\z|\x, m)$. 

In many cases, the solution output is placed at a two-dimensional grid,  $\y_m(\x) \in \mathbb{R}^{d_1 \times d_2}$, \eg a 1D spatial-temporal grid or a 2D spatial grid. We therefore can view the solution output and the SDE state $\z(t)$ as images. We use the U-Net architecture~\citep{ronneberger2015u} to design our conditional score model $\s_\btheta(\z, t, \x, \m)$. U-Net was first developed for image segmentation and then has been broadly used in deep generative models for image generation~\citep{ho2020denoising,song2019generative,song2021scorebased}. In general, U-Net consists of a contracting path that repeatedly reduces the image resolution (but adding more and more channels), and then an expansion path that recovers the resolution step by step. \cmt{Similar to~\citep{song2021scorebased}, }We use two ResNet~\citep{he2016deep}  blocks followed by a down-sampling or up-sampling operation to implement each step in the two paths (see the details in the Appendix). To incorporate the PDE parameters $\x$ and fidelity $m$, we first obtain an embedding for each of them, 
\begin{align}
	\e_1(\x) = \text{MLP}(\x), \;\; \e_2(m) = \text{MLP}(m) \label{eq:embedding}
\end{align}
where $\text{MLP}$ means multi-layer-perception. We then integrate the embeddings into the output of the first convolution layer in each ResNet block $k$, 
\begin{align}
	\Hcal_k  \leftarrow \Hcal_k + \sigma(\W_{1k} \e_1) + \sigma(\W_{2k} \e_2) \label{eq:adding-to-resnet}
\end{align}
where $\Hcal_k$ is the $r_k \times h_k \times c_k$ output, $c_k$ is the number of channels, the resolution is $r_k \times h_k$, $W_{1k}$ is of size $c_k \times s_1$ and $W_{2k}$ of size $c_k \times s_2$,  $s_1$ and $s_2$ are the dimensions of $\e_1$ and $\e_2$, respectively, and $\sigma(\cdot)$ is the activation.  The SDE time $t$ is incorporated in a similar way, but we construct a random Fourier feature (RFF) vector to represent $t$, which is consistent with~\citep{song2021scorebased}. In this way, the full conditional information is integrated into every unit of the U-Net to estimate the conditional score function $\nabla_{\z} \log q_t(\z|\x, m)$. Our model is illustrated in Fig. \ref{fig:score-model-framework}. 

However, when we predict the solution of more complex PDEs, the output can be placed at a higher dimensional grid, for which we cannot directly apply our U-Net based model. For example, to solve a 2D diffusion equation in a time range,  the solution output is placed in a three-dimensional tensor, $y_m(\x) \in \mathbb{R}^{d_1 \times d_2 \times d_3}$, where $d_1$ and $d_2$ correspond to two spatial indexes and $d_3$ is the number of time steps. To predict the dynamics for a 3D spacial temporal problem, the solution output becomes a four-dimensional tensor, since we have three spacial indexes. Extending U-Net to higher dimensional data is not easy and can dramatically increase the model parameters and training challenge. To sidestep this issue, we propose a simple and effective approach. Take the three-dimensional solution array (tensor) as an example. We consider generating each slice of the solution at a specific time step $\tau$,  $\y_m(\x, \tau) \in \mathbb{R}^{d_1 \times d_2} \sim p(\cdot | \x, m, \tau)$. Hence, we extend the diffusion-generation model in \eqref{eq:forward-sde} and \eqref{eq:backward}, to be conditioned not only on $\x$ and $m$ , but also on $\tau$.  We use a model $\s_\btheta(\z, t, \x, m, \tau)$ to estimate the conditional score  $\nabla_{\z} \log q_t(\z|\x, m, \tau)$. Since each solution slice is viewed as an image, we can still use our U-net based architecture. To incorporate the time step $\tau$, we follow \eqref{eq:embedding} and \eqref{eq:adding-to-resnet} to first obtain an embedding via MLP and then add it into the output channels of the first convolution layer in each block.  Similarly, for the 3D spatial temporal solution output (which is the highest dimension in most practical problems), we can estimate a score model conditioned on $\x$, $m$,  $\tau$ and one spacial index. In this way, we still generate 2D solution slices. 

One advantage of our method is that it naturally unifies modeling with discrete and continuous fidelities. When $m$ is discrete, the input to the embedding network in \eqref{eq:embedding} is a one-hot encoding. When $m$ is continuous, we feed the continuous fidelity value outright to the network to obtain the embedding. 

\subsection{Algorithm}
To estimate the model parameters $\btheta$, we  use the denoising score matching framework~\citep{song2019generative}, 
\begin{align}
	\btheta^* &= \arg \min_\btheta \EE_t  \EE_{\x, m}\EE_{\z(0)|\x, m} \EE_{\z(t)|\z(0)}[\Lcal_t], \notag \\
	\Lcal_t&= \left\|\s_\btheta\left(\z(t), t, \x, m\right) - \nabla_\z \log q\left(\z(t)|\z(0)\right)\right\|^2,
\end{align}
where $q(\z(t)|\z(0))$ is the distribution of the state $\z(t)$ conditioned on the initial state $\z(0)$ in the forward SDE. 
While the training objective does not have an analytical form, we can apply a stochastic optimization. Specifically, given a large $T$ (at which we believe $q_T$ has approximately become a Gaussian white noise distribution), we sample a mini-batch of $t$ from $[0, T]$ and $\{\x, m, \z(0)\}$ from the training dataset $\Dcal$. Due to the simple form chosen for the forward SDE \eqref{eq:forward-sde},  it is easy to obtain $q\left(\z(t)|\z(0)\right)$. For example, if using the VE-SDE, we have $q(\z(t)|\z(0)) = \N\left(\z(t)|\z(0), (v(t) - v(t))\I\right)$. Next, we sample $\z(t)$ from $q(\z(t)|\z(0))$ to compute $\s_\btheta(\z(t), t, \x, m)$ and  $\Lcal_t$. 
We can use automatic differentiation libraries to compute the stochastic gradient $\nabla_{\btheta} \Lcal_t$ with which to update $\btheta$. 
Note that with the VE-SDE, we have $$\Lcal_t = \left\|\s_\btheta\left(\z(t), t, \x, m\right) - \frac{\z(0) - \z(t)}{v(t) - v(0)}\right\|^2.$$ Since $\z(0) - \z(t)$ is a random Gaussian noise, minimizing (expected) $\Lcal_t$ is essentially to fit the noise. \cmt{From other commonly used diffusion SDEs~\citep{song2021scorebased}, we can observe the same conclusion.} This exhibits the key difference from the likelihood-based approaches, which mainly fit the data $\Dcal$ rather than the noise. 

For prediction, suppose we are given new PDE parameters $\x^*$ and the target fidelity $m^*$. We first sample a noise $\z(T)$ from $q_T$\cmt{  (for VE SDE, it corresponds to a zero-mean Gaussian distribution with a large variance)} as the initial state of the backward SDE \eqref{eq:backward}. Using the learned conditional score function $s_\btheta(\z(t), t, \x^*, m^*)\approx \nabla_{\z} \log q_t(\z|\x^*, m^*) $, we then solve (simulate) the backward SDE to time $0$ and obtain the prediction $\z(0) \overset{\Delta}{=} \y_{m^*}(\x^*)$.  

\section{Related Work}
Linear model of coregionalization (LMC)~\citep{matheron1982pour, goulard1992linear} is perhaps the most popular high-dimensional output regression model, which has many variants, such as ~\citep{goovaerts1997geostatistics,higdon2008computer,xing2016manifold,xing2015reduced}. Other multi-output GP regression models include~\citep{higdon2002space,boyle2005dependent,alvarez2019non,bonilla2007kernel,rakitsch2013all,wilson2012gaussian,zhe2019scalable}. \etc But most of these methods do not scale to large numbers of outputs.

 \citet{perdikaris2017nonlinear,cutajar2019deep} first used a nonlinear auto-regressive framework to implement multi-fidelity learning of single-output GP models. To handle high-dimensional outputs, \citet{wang2021multi} used a similar architecture on the LMC, and placed a matrix GP prior to connect successive fidelities. \citet{li2020deep} developed a Bayesian auto-regressive neural network and an active learning algorithm for multi-fidelity high-dimensional output learning. For the same model, \citet{li2022batch} developed a batch active learning algorithm with budget constraints. \citet{li2022infinite} proposed a continuous-fidelity high-dimensional surrogate learning method. Other recent multi-fidelity models include \citep{hamelijnck2019multi,wang2020mfpc,wu2022multi,xing2021residual}, \etc

There have been proposed many generative models, such as  likelihood-based approaches, including auto-regressive models~\citep{larochelle2011neural,germain2015made,van2016pixel},  normalization flows~\citep{dinh2014nice,dinh2016density,papamakarios2017masked}, variational auto-encoders~\citep{kingma2013auto,rezende2014stochastic}, energy-based models~\citep{lecun2006tutorial,gutmann2010noise,song2021train}, \etc A second class is implicit generative models~\citep{mohamed2016learning}, \eg Generative Adversarial Networks (GAN)~\citep{goodfellow2014generative,salimans2016improved,arjovsky2017wasserstein}. The recent diffusion-generative models (also called score-based models) use a denoising process to generate the samples, including  denoising diffusion probabilistic models (DDPM)~\citep{ho2020denoising}, score matching with Langevin dynamics~\citep{song2019generative}, score-based models through SDEs~\citep{song2021scorebased}, \etc. They have outperformed other types of methods and achieved the new state-of-the-art in many tasks, such as in generating images~\citep{ho2020denoising}, audio~\citep{kong2020diffwave,chen2020wavegrad}, and graphs~\citep{niu2020permutation}. 

\section{Experiment}
\subsection{Predicting PDE Solution Fields}
We first evaluated \ours in predicting the solution fields of three benchmark PDEs: Heat, Poisson's, and Burgers' equations. 

{\bf Datasets.} Training and test examples were collected by solving these PDEs with different meshes. 
The denser the mesh, the higher the fidelity. For Heat and Poisson's equations, we used four fidelities, corresponding to meshes of $8 \times 8$, $16 \times 16$, $32 \times 32$ and $64 \times 64$, respectively, and  for Burgers' equation, we used  $16 \times 16$, $24 \times 24$, $32 \times 32$ and $64 \times 64$ meshes. For each PDE, 
the number of training  examples at each fidelity (from low to high) is 128, 64, 32, and 8, respectively.
For testing, we computed $128$ examples at the highest fidelity. The training/test inputs are the PDE parameters uniformly sampled from a specific domain. We followed the details given in~\citep{wang2021multi}. 

{\bf Competing Methods.} We compared with following state-of-the-art multi-fidelity learning methods for high-dimensional output regression. (1) DRC~\citep{xing2021deep}, deep residual coregionalization, which learns an LMC to predict the residual error between successive fidelities. The final prediction is made by adding the predictions of models across the fidelities. (2) DMF~\citep{li2020deep}, deep multi-fidelity learning via a Bayesian auto-regressive neural network. (3) MFHoGP~\citep{wang2021multi}, a nonparametric multi-fidelity extension of LMC. (4) IFC-ODE$^2$ and (5) IFC-GPODE~\citep{li2022infinite}, infinite fidelity coregionalization models, which use a neural ODE~\citep{chen2018neural} to model  how a low-dimensional solution representation varies with the continuous fidelity, and predict the high-dimensional solution output by multiplying with a basis matrix. To estimate a fidelity-varying basis matrix, IFC-ODE$^2$ uses another element-wise neural ODE  while IFC-GPODE uses a GP prior over each basis element. We also tested a plain (6) LMC, which incorporates all the training examples without differentiating their fidelities. Except DMF, all the methods use interpolation~\citep{zienkiewicz1977finite} to align the solution output of every example to the same gird as in the highest fidelity. 

We tested our method with both discrete and continuous fidelity views. For the discrete case, the fidelity input to our conditional score model (see \eqref{eq:embedding}) is a one-hot encoding. For the continuous case, we followed~\citep{li2022infinite} to map the fidelity value to $[0, 1]$. Specifically, the fidelity of an $s \times s$ mesh is computed as $m(s) = (s - s_0)/(s_1 - s_0)$ where $s_0 \times s_0$ is the mesh size for the lowest fidelity in the training data, and $s_1 \times s_1$ the highest fidelity. We denote our method with discrete fidelities by \ours-D and with continuous fidelities by \ours-C. To confirm that our method indeed utilizes low-fidelity data to improve high-fidelity predictions, we also examined our method only using the high-fidelity examples, denoted by \ours-HF-OLY.  


{\bf Settings and Results.} We implemented \ours with Pytorch~\citep{paszke2019pytorch}, LMC with Matlab, and used the original implementation of the other methods. For \ours, we used a two-layer MLP, with ten neurons per layer, to compute the embeddings for the PDE parameters $\x$ and fidelity $m$ (see \eqref{eq:embedding}). For the SDE time $t$, we generated 64 random Fourier features. We used the Swish activation. We used the Variance Exploding (VE) SDE as the forward SDE.  For solution generation, we used the Predictor-Corrector (PC) sampler~\citep{song2021scorebased}, where the number of sampling steps was set to 2,000, and Langevin dynamics steps was set to 1. Since the solution generation is a random sampling procedure, we tested the performance of our method with only one sampled solution and with the average of five sampled solutions. 
For training, DRC uses L-BFGS while all the other methods uses ADAM~\citep{kingma2014adam}. We set the learning rate to $2\times10^{-4}$ for \ours and $10^{-3}$ (the default choice) for other methods. For IFC-ODE$^2$ and IFC-GPODE, we used the Runge-Kutta method of order 5 with adaptive steps as the ODE solver. We used the square exponential (SE) kernel in DRC, MFHoGP, and IFC-GPODE.\cmt{, and initialized the length-scale parameter as 1. These are all the default settings in the original implement ion of the methods.}  We set the dimension of the latent output or the number of bases to 20 in the competing methods. For DMF, we followed~\citep{li2020deep} to use two hidden layers for the NN in each fidelity, and $\mathrm{tanh}$ activation. The number of neurons per layer was chosen from $\{20, 30, 40, 50\}$. We repeated the evaluation five times, and each time we re-generated a new training dataset and test dataset. We report the average relative $L_2$ error and the standard deviation of each method  in Table \ref{tb:l2-pdes}.

\begin{table}
		\centering
		\small
		\begin{tabular}[c]{cccc}
			\hline\hline
			Methods &  \multicolumn{3}{c}{Error Reduction \%}  \\
			\cmidrule(l){2-4}
			& \textit{Poisson's}  & \textit{Heat} & Burger's \\
			\hline 
			\ours: Best   & $55.6$ & $58.0$& $45.4$ \\
			\ours: 2nd Best   & $42.1$	& $53.1$ & $44.5$ \\
			\hline
		\end{tabular}
	\caption{\small  Error reduction percentage as compared with the top-performed competing approach.}
	\label{tb:err-reduct}
\end{table}
As we can see, \ours, when using the full training dataset,   
 consistently outperforms all the competing approaches by a large margin (under all the settings). The percentage of the error reduction in contrast to the top-performed competing method is shown in Table \ref{tb:err-reduct}.  \ours tremendously improves upon the competing baselines. Even the second best result of \ours can reduce the relative error of the top-performed baseline by more than 40\%. Overall, it is better to average multiple solution samples for prediction than just use one. Although generating multiple samples is more expensive, these samples can be generated completely in parallel, and hence it does not hurt the prediction efficiency. In addition, we can see that when \ours only uses the high-fidelity examples (\ie \ours-D-HF-OLY and \ours-C-HF-OLY), the predictive performance drops significantly. It, therefore, shows that \ours can indeed leverage lower-fidelity data to bolster the predictive performance at the high-fidelity. 
The relative $L_2$ errors of LCM and DRC are much bigger than the others for Poisson's equation, which implies both methods failed. This might be due to the relatively big scale difference between the low-fidelity and high-fidelity training outputs. Since the low-fidelity examples are much more than the high-fidelity examples,  the bases computed from PCA might be dominated by the low-fidelity data. Together these results have shown a great advantage of the proposed diffusion-generative model. 
 
\begin{table*}
	\begin{subtable}{\textwidth}
     \centering
	\small
	\begin{tabular}[c]{cccc}
		\hline\hline
		\centering
		Method &  \multicolumn{3}{c}{Relative $L_2$ Error}  \\
		\cmidrule(l){2-4}
		& \textit{Poisson's}  & \textit{Heat } & \textit{Burgers'} \\
		\hline 
		LMC   & $65.1048 \pm	0.5259$ & $0.0865	\pm 0.0003$ &  $0.0444 \pm	0.0024$\\
		DRC         &  $26.4498 \pm	0.6392$ & $0.0244 \pm	0.0005$& $0.0317 \pm	0.0004$\\
		DMF        &  $0.2152 \pm	0.0040$ & $0.0390 \pm	0.0033$ & $0.0438\pm	0.0088$ \\
		MFHoGP  & $0.4031 \pm	0.2410$ & $0.0138	\pm 0.0029$& $0.0349\pm0.0077$ \\
		IFC-ODE$^2$   & $0.4690 \pm	0.2860$ & $0.0112 \pm	0.0015$ & $0.0365 \pm	0.0050$ \\
		IFC-GPODE  & $0.2097	 \pm 0.0498$ & $0.0081 \pm	0.0014$ & $0.0331 \pm	0.0045$ \\
		\hline
		\ours-D-HF-OLY & $0.1801	 \pm 0.0091$ & $0.1160 \pm	0.0128$ & $0.0584 \pm	0.0120$\\
		\ours-C-HF-OLY & $0.2431 \pm	0.0474$ & $ 0.1049 \pm	0.0132$ & $0.0572 \pm	0.0116$ \\
		\ours-D-1 & $0.1713\pm	0.0226$ & $0.0038 \pm	0.0002$ & $0.0176\pm	0.0032$\\
		\ours-D-5 & $0.1454 \pm	0.0232$ & ${\bf 0.0034 \pm	0.0003}$ & ${\bf 0.0173 \pm	0.0037}$ \\
		\ours-C-1 & $0.1214\pm	0.0074$ & $0.0040 \pm	0.0006$ & $0.0217\pm	0.0041$ \\
		\ours-C-5 & ${\bf0.0932\pm	0.0045} $ & $0.0038 \pm	0.0006$ & $0.0218 \pm 0.0042$ \\
		\hline
	\end{tabular}
	\end{subtable}
\caption{\small Average relative $L_2$ error for predicting solution fields of Poisson's, Heat and Burgers' equations. The results were averaged from five runs. ``-D'' and ``-C'' denote the discrete and continuous fidelities, respectively. ``-1'' and ``-5'' mean using one sampled solution and the average of five sampled solutions for prediction. ``-HF-OLY'' means training only with examples at the highest fidelity. }
\label{tb:l2-pdes}
\end{table*}
\begin{table}
	\centering
	\small
	\begin{tabular}[c]{cc}
		\hline\hline
		Method &  \multicolumn{1}{c}{Relative $L_2$ Error}  \\
		\hline 
		LMC   & $0.4736	\pm 0.0040$\\
		DRC         & $0.4864	\pm 0.0042$ \\
		DMF        &  $0.4192	\pm 0.0088$\\
		MFHoGP  & $0.4099	\pm 0.0139$ \\
		IFC-ODE$^2$   & $0.4209 \pm	0.0101$ \\
		IFC-GPODE  & $0.3759 \pm	0.0172$ \\
		\hline
		\ours-D-1 & $0.3408	\pm 0.0148$ \\
		\ours-D-5 & $0.3284 \pm	0.0124$  \\
		\ours-C-1 &  $0.3198\pm	0.0185$ \\
		\ours-C-5 &  ${\bf 0.3079 \pm	0.0186}$ \\
		\hline
	\end{tabular}
	\caption{\small Average relative $L_2$ error and the standard deviation for predicting optimal topological structures. The results were averaged over five runs.}
	\label{tb:l2-topopt}
\end{table}
\begin{table}
	\centering
	\small
	\begin{tabular}[c]{ccc}
		\hline\hline
		Method &  \multicolumn{2}{c}{Relative $L_2$ Error}  \\
		\cmidrule(l){2-3}
		& \textit{NS-2} & \textit{NS-20}\\ 
		\hline 
		LMC   & $0.7571	\pm 0.0111$ & $0.7919 \pm	0.0146$\\
		DRC         &$0.3241 \pm	0.0081$ & $0.4107 \pm	0.0225$\\
		DMF        &  $0.2945	\pm 0.0218$ & $0.4912 \pm	0.0188$ \\
		MFHoGP  & $0.1147\pm	0.0100$ & $0.3703 \pm	0.0193$  \\
		IFC-ODE$^2$   & $0.0899 \pm 0.0052$  & N/A\\
		IFC-GPODE  & $0.0934	 \pm 0.0053$ & N/A \\
		\hline
		\ours-D-1 & $0.0810 \pm	0.0109$ & $0.2539 \pm	0.0110$ \\
		\ours-D-5 & $0.0774 \pm	0.0113$  & $0.2333 \pm	0.0133$ \\
		\ours-C-1 & $0.0767 \pm	0.0122$  & $0.2148 \pm	0.0272$  \\
		\ours-C-5 & ${\bf 0.0710 \pm	0.0118}$ & ${\bf 0.1939 \pm	0.0279}$ \\
		\hline
	\end{tabular}
	\caption{\small Average relative $L_2$ error for predicting pressure fields in fluid dynamics.  \textit{NS-2} means predicting two slices of the pressure field solved by the Navier-Stokes (NS) equation at time $\tau=0$ and $\tau=10$. \textit{NS-20} means predicting 20 slices of the pressure field at $\tau \in [0:0.5:10]$.  The results were averaged from five runs.}
	\label{tb:l2-ns}
\end{table}
\subsection{Topology Optimization}
Next, we applied \ours in predicting the optimal topology structures given different design parameters. Topology optimization (TO)  is the key step in many design tasks.
To learn a surrogate model, we can view the design parameters as the input, and the corresponding optimal structure as the output. Topology optimization is often computationally very expensive. Numerical solvers are often used to obtain the solution field of relevant PDEs as an intermediate step. Quantities of interest are computed and optimized based on the solution field.  Hence, the mesh or finite element size can determine the fidelity of the final structure. We followed~\citep{keshavarzzadeh2019parametric} to consider the design of an L-shape linear elastic structure at $[0, 1] \times [0, 1]$ such that given a load on the bottom right half (the location is in $[0.5,1]$ and angle in $[0, \frac{\pi}{2}]$), the structure achieves the maximum stiffness. 

We used four fidelities to collect training structures at different load parameters (\ie input), which correspond to applying $16 \times 16$, $32 \times 32$, $64 \times 64$ and $128 \times 128$ meshes in the internal solver. The numbers of examples are $128$, $64$, $32$ and $8$, respectively. Another $128$ examples at the highest fidelity were generated for testing. We repeated the evaluation five times, and report the average relative $L_2$ error of each method and the standard deviation in Table \ref{tb:l2-topopt}. 

We can see that \ours consistently outperforms all the competing methods by a large margin. The best performance is obtained when we used the continuous fidelity value and an average of five solution samples for prediction, \ie \ours-C-5. It reduces the error of LMC, DRC, DMF, MFHoGP, IFC-ODE$^2$ and IFC-GPODE by 35.0\%, 36.7\%, 26.6\%, 24.9\%,  26.8\%, 18.1\%, respectively. The results demonstrate the great superiority of \ours in prediction accuracy for topology optimization.  
\begin{figure*}
	\centering
	\setlength\tabcolsep{0pt}
	\begin{tabular}[c]{cc}
		\setcounter{subfigure}{0}
		\begin{subfigure}[t]{0.5\textwidth}
			\centering
			\includegraphics[width=\textwidth]{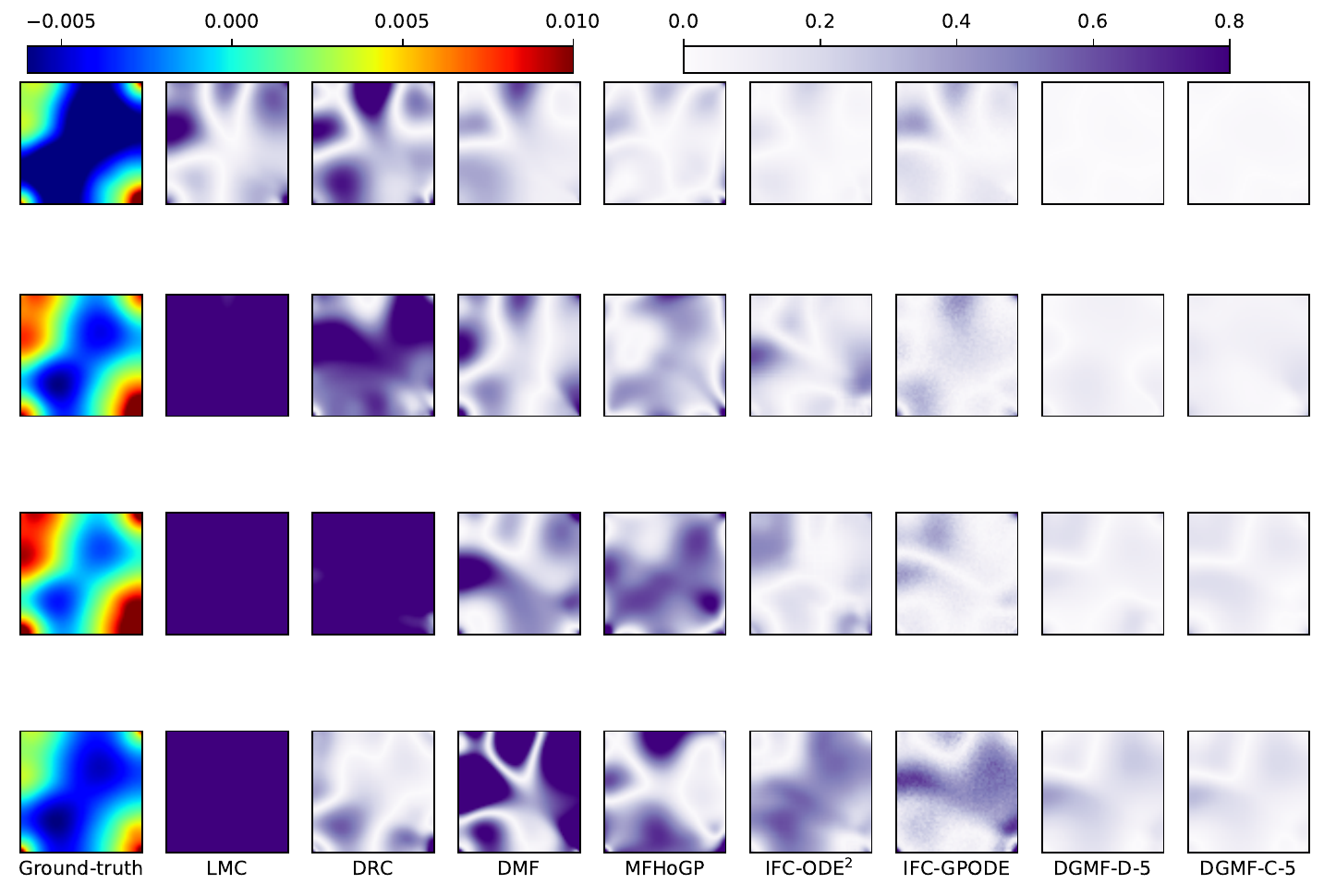}
			\caption{\small Pressure field at several time points.}
		\end{subfigure} 
		&
		\begin{subfigure}[t]{0.5\textwidth}
			\centering
			\includegraphics[width=\textwidth]{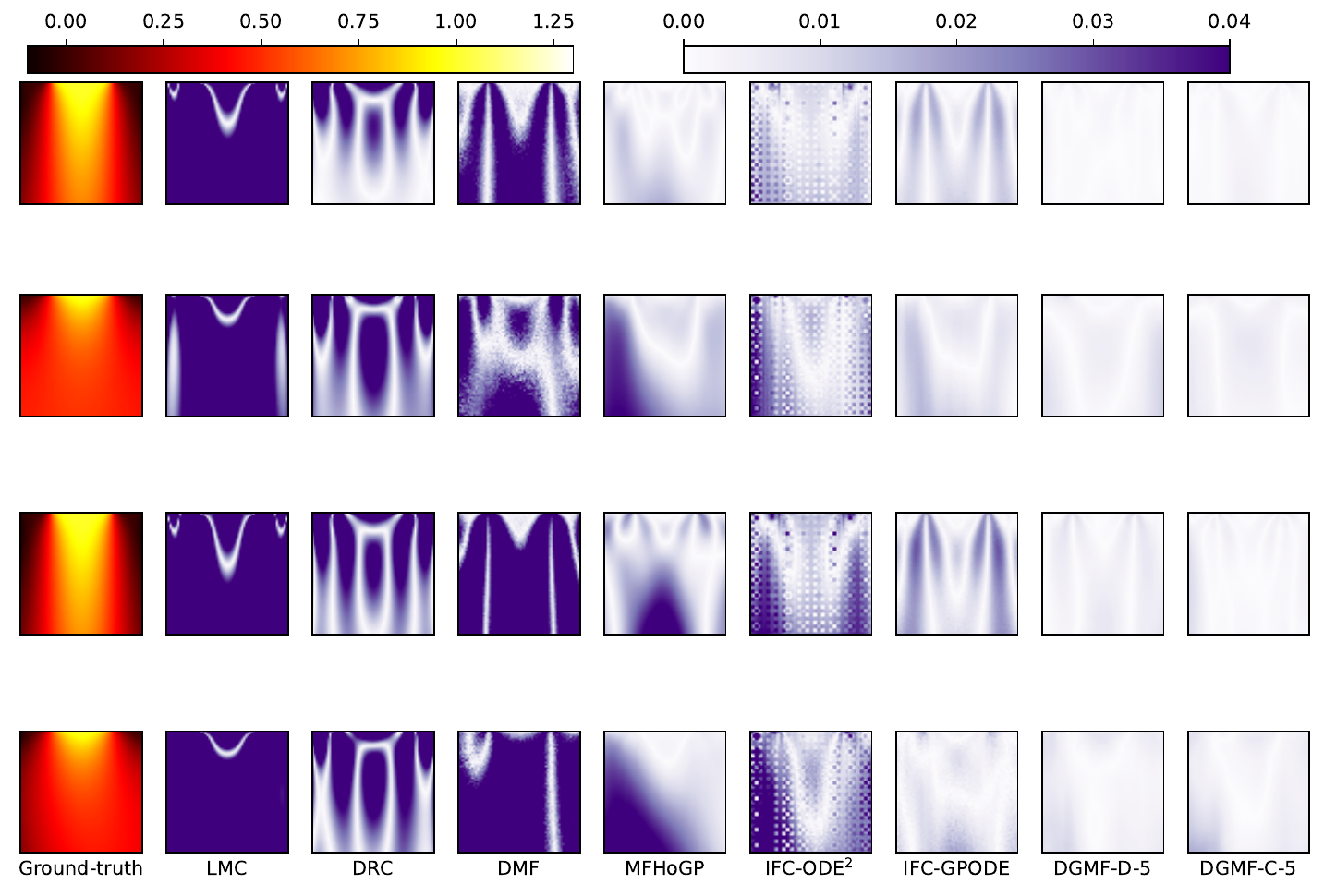}
			\caption{\small \textit{Heat} equation}
		\end{subfigure}
	\end{tabular}
	\vspace{-0.1in}
	\caption{\small Individual output prediction errors. The original solutions are shown in the leftmost columns of (a) and (b). Each other column shows the normalized absolute error of one method. Lighter colors indicate smaller errors. \cmt{ of the of the prediction error field. For NS, we only display the pressure field when t=10.0(the last episode). All the error fields of compared methods are normalized to the same scale $[0.0, \max(y_i)]$}} \label{fig:error_field}
	\vspace{-0.1in}
\end{figure*}

\subsection{Computational Fluid Dynamics}
\vspace{-0.05in}
Third, we evaluated \ours in a computational fluid dynamics problem. Following~\citep{li2022infinite}, we considered a flow in a rectangular domain $[0, 1] \times [0, 1]$, which is driven by the boundaries with a prescribed tangential velocity at each boundary. The flow dynamics can be computed by solving the incompressible Navier-Stokes (NS) equations~\citep{chorin1968numerical}, which is known to be computationally expensive with large Renolds numbers. We intended to predict the pressure field of the flow with time $\tau \in [0, 10]$, given the four boundary velocities, each  in $[0, 10]$, and the Reynolds number in $[10, 1000]$. We collected training examples at five fidelities, corresponding to spatial meshes of $32 \times 32$, $64 \times 64$, $96 \times 96$, $112 \times 112$, and $128 \times 128$. The time step was  set to $0.05$. We collected 128, 64, 32, 16 and 8 examples for the five fidelities (from low to high). Another $128$ examples at the highest fidelity were collected for testing. We considered two tasks. The first task is to predict the pressure field at $t=0$ and $t=10$, namely, 2 slices. The second task is to predict the pressure field at $20$ time points, $t \in [0:0.5:10]$, \ie 20 slices. For both tasks, the training and test outputs are three-dimensional tensors, \ie $128 \times 128 \times 2$ and $128 \times 128 \times 20$. In our method, we, therefore, introduced the time $\tau$ as an additional input to the conditional score model to generate the pressure field at  $\tau$, which can still be viewed as an image. We show the average $L_2$ relative error of each method from five independent experimental trials in Table \ref{tb:l2-ns}. Note that for the second task, since the training output dimension is very large (328K), both IFC-ODE$^2$, and IFC-GPODE ran out of memory and their results are not available\footnote{The original implementation of IFC uses automatic differentiation to back-propagate the gradient during ODE solving. The memory usage quickly exploded due to the growth of the computational graph. We also tried with adjoint-state approach~\citep{chen2018neural}, which however rendered the training too slow to obtain meaningful results.  }.

From Table \ref{tb:l2-ns}, we can see that \ours again outperforms all the competing methods, which is consistent with the other experiments. \ours can reduce the error of the best-performed baseline by 21\% and 47.6\% in task 1 and 2, respectively.  
While  IFC-ODE$^2$ and IFC-GPODE achieved significantly smaller prediction errors than the other competing methods in the first task (predicting 2 time slices), they are much more expensive in computation and memory consumption, and cannot scale to the second task (predicting 20 time slices). Note that for most methods, the accuracy of predicting 20 slices is worse than predicting 2 slices. This is reasonable because the former is much more challenging, yet both tasks used the same number of training examples (at each fidelity).

We further investigated the prediction error of individual solution outputs. We randomly selected four test examples for fluid dynamics (at different time points) and the Heat equation, and examined the normalized absolute error\footnote{it is normalized by the average output (pixel) value.} of each method in predicting every single output. The results are visualized in Fig. \ref{fig:error_field}. As we can see, these cases appear to be more challenging to the competing approaches. There are many places with dominant local errors, implying inferior prediction of the solution at these places. By contrast, \ours exhibit much smaller and more uniform local errors, which leads to a superior global error.

\textbf{Computational Efficiency.} Finally, we examined the computational efficiency of \ours. We tested all the methods on a Ubuntu server, with one NVIDIA A100-SXM4-40GB GPU and an AMD EPYC 7J13 64-Core Processor. The memory is 200G. We used Pytorch 1.13.1 and Cuda 11.7.  The training time per-iteration/-epoch is reported in Table \ref{tb:running-time}. As we can see, the training efficiency of \ours is comparable to the existing multi-fidelity approaches. However, a drawback of our method is that the solution generation is much slower than the other methods. This is a well-known problem of the diffusion-denoising framework.  However, compared with numerical solvers, our method is still much faster. For example, in computational fluid dynamics,  the average time for \ours to predict a solution slice ($128 \times 128$) at $\tau=10$ is 1.17 seconds, and in topology optimization, the average time of \ours to predict a structure is 1.11 seconds. By contrast, running numerical solvers for the two tasks takes 360 and 200 seconds on average. Our method still achieves  300x and 180x speed-ups, respectively.

\begin{table}
	\centering
	\small
	\begin{tabular}[c]{c|cc}
		\hline\hline
		Method &  \multicolumn{2}{c}{Training time per-epoch (seconds)}  \\
		\cmidrule(l){2-3}
		& Topology optimization & NS-2\\
		\hline 
		LMC   & 0.02 & 0.03\\
		DRC   & 0.04 & 0.05 \\
		DMF   & 0.08 &  0.82\\
		MFHoGP & 0.06 & 0.17 \\
		IFC-ODE$^2$   &1.59 &  0.57 \\
		IFC-GPODE  & 7.45 & 1.34 \\
		\ours & 0.20 & 0.22 \\
		\hline
	\end{tabular}
	\vspace{-0.1in}
	\caption{\small Average per-epoch/-iteration training time.}
	\label{tb:running-time}
	\vspace{-0.2in}
\end{table}

\vspace{-0.05in}
\section{Conclusion}
We have proposed \ours, a novel diffusion-generative multi-fidelity learning method for physical simulation. Different from the existing methods, \ours modeled the PDE solution as gradually generated from random noises, and the generation is controlled by the PDE parameters and fidelity. The initial results have shown a great advantage in prediction accuracy. In the future work, we will develop more efficient prediction/generation algorithms and investigate more applications in computational physics.

\bibliographystyle{apalike}
\bibliography{MFDiffGen}
%
%

\newpage
\appendix
\section{Appendix}
\subsection{U-Net Backbone Details}
Similar to \citep{brock2018large}, we use ResNet blocks to construct the contracting and expansion path in the U-Net. That is, each time we feed the output at the previous resolution to 2 consecutive ResNet blocks and then apply a down-sampling or up-sampling operation (via a Fir filter) to reduce or increase the resolution by 2.  At the resolution $16 \times 16$, we apply a channel-wise self-attention over the output of the second ResNet block. Each ResNet block is fulfilled by repeatedly applying group normalization, convolution with padding, and nonlinear activation twice, and then adding the skip layer. At the bottleneck (where we have the smallest resolution $4 \times 4$), we apply a ResNet block,  a channel-wise self-attention, and then a ResNet block. For the input solution (image) of size $64 \times 64$, the number of channels at each resolution (reduced by 2 each time) is [16, 16, 32, 32, 64], and for the solution of size $128 \times 128$, is [16, 16, 32, 32, 64, 64].

\end{document}